\documentclass[sigconf]{acmart}

\AtBeginDocument{%
  \providecommand\BibTeX{{%
    \normalfont B\kern-0.5em{\scshape i\kern-0.25em b}\kern-0.8em\TeX}}}

\copyrightyear{2022}
\acmYear{2022}
\setcopyright{acmlicensed}
\acmConference[KDD '22]{Proceedings of the 28th ACM SIGKDD Conference on Knowledge Discovery and Data Mining}{August 14--18, 2022}{Washington, DC, USA}
\acmBooktitle{Proceedings of the 28th ACM SIGKDD Conference on Knowledge Discovery and Data Mining (KDD '22), August 14--18, 2022, Washington, DC, USA}
\acmPrice{15.00}
\acmDOI{10.1145/3534678.3539444}
\acmISBN{978-1-4503-9385-0/22/08}

\usepackage{import}
\usepackage{./highlight}

\usepackage{graphicx} %
\usepackage{amsmath}
\usepackage{booktabs}
\usepackage{subcaption}
\usepackage{amsthm}
\usepackage{algorithm}
\usepackage[noend]{algpseudocode}
\usepackage{multirow}

\usepackage{adjustbox}
\usepackage{array}

\usepackage{systeme,mathtools}
\usepackage[textsize=footnotesize]{todonotes}

\usepackage{bbm}

\usepackage{tikz}
\usetikzlibrary{shapes,decorations,arrows,calc,arrows.meta,fit,positioning}
\tikzset{
    -Latex,auto,node distance =1 cm and 1 cm,semithick,
    state/.style ={circle, draw, minimum width = 1.0 cm,inner sep=0pt},
    point/.style = {circle, draw, inner sep=0.04cm,fill,node contents={}},
    bidirected/.style={Latex-Latex,dashed},
    el/.style = {inner sep=2pt, align=left, sloped}
}

\newcommand*{\missingreference}{{\Huge \colorbox{red}{?reference?}}}
\newcommand*{\missingcitation}{{\Huge \colorbox{red}{?citation?}}}
\makeatletter
\def\@setref#1#2#3{%
  \ifx#1\relax
    \protect\G@refundefinedtrue
    \nfss@text{\reset@font\missingreference}%
    \@latex@warning{Reference `#3' on page \thepage \space
      undefined}%
  \else
    \expandafter#2#1\null
  \fi}
\def\@citex[#1]#2{\leavevmode
  \let\@citea\@empty
  \@cite{\@for\@citeb:=#2\do
    {\@citea\def\@citea{,\penalty\@m\ }%
      \edef\@citeb{\expandafter\@firstofone\@citeb\@empty}%
      \if@filesw\immediate\write\@auxout{\string\citation{\@citeb}}\fi
      \@ifundefined{b@\@citeb}{\hbox{\reset@font\missingcitation}%
        \G@refundefinedtrue
        \@latex@warning
        {Citation `\@citeb' on page \thepage \space undefined}}%
      {\@cite@ofmt{\csname b@\@citeb\endcsname}}}}{#1}}
\makeatother

\newtheorem{problem}{Problem}
\newtheoremstyle{case}{}{}{}{}{}{:}{ }{}
\theoremstyle{case}

\newcommand{\features}{\mathbf{X}}
\newcommand{\remove}{\mathbf{K}}

\DeclareMathOperator*{\argmin}{arg\,min\,\,}

\newcommand{\bftab}{\fontseries{b}\selectfont}

\newcolumntype{R}[2]{%
    >{\adjustbox{angle=#1,lap=\width-(#2)}\bgroup}%
    l%
    <{\egroup}%
}
\newcolumntype{P}[1]{>{\arraybackslash}p{#1}}
\newcolumntype{M}[1]{>{\centering\arraybackslash}m{#1}}
\newcolumntype{N}{@{}m{0pt}@{}}

\newcommand{\subsetname}{causal feature set}

\newcommand{\Combined}{Heterogeneous Treatment Effect Feature Selection (HTE-FS)}
\newcommand{\combined}{HTE-FS}

\newcommand{\Globalmethodname}{HTE-Fit}
\newcommand{\globalmethodname}{HTE-Fit}
\newcommand{\Globalmethodnamenoparen}{HTE-Fit}

\newcommand{\Localmethodname}{Structure-Fit}
\newcommand{\localmethodname}{Structure-Fit}
\newcommand{\Localmethodnamenoparen}{Structure-Fit}

\newcommand{\Globalforward}{\globalmethodname{} --- Forward}
\newcommand{\Globalbackward}{\globalmethodname{} --- Backward}
\newcommand{\globalforward}{\globalmethodname{}-F}
\newcommand{\globalbackward}{\globalmethodname{}-B}

\newcommand{\Fullstructure}{Full Structure Learning (FSL)}
\newcommand{\fullstructure}{FSL}

\newcommand{\new}[1]{#1}

\newcommand{\kdd}[1]{{\textcolor{blue}{\textit{#1}}}}
\renewcommand{\kdd}[1]{#1}

\def\true{true}
\def\arxiv{false}

\ifx\arxiv\true
\newcommand{\yesappendix}[1]{}
\newcommand{\noappendix}[1]{{\textcolor{blue}{\textit{#1}}}}
\else
\newcommand{\yesappendix}[1]{#1}
\newcommand{\noappendix}[1]{} 
\fi

\begin{document}

\title{Improving Data-driven Heterogeneous Treatment Effect Estimation Under Structure Uncertainty}

\author{Christopher Tran}
\affiliation{%
	\institution{University of Illinois at Chicago}
	\city{Chicago}
	\country{IL, USA}
}
\email{ctran29@uic.edu}

\author{Elena Zheleva}
\affiliation{%
	\institution{University of Illinois at Chicago}
	\city{Chicago}
	\country{IL, USA}
}
\email{ezheleva@uic.edu}

\begin{abstract}
Estimating how a treatment affects units individually, known as heterogeneous treatment effect (HTE) estimation, is an essential part of decision-making and policy implementation. The accumulation of large amounts of data in many domains, such as healthcare and e-commerce, has led to increased interest in developing data-driven algorithms for estimating heterogeneous effects from observational and experimental data. However, these methods often make strong assumptions about the observed features and ignore the underlying causal model structure, which can lead to biased HTE estimation. At the same time, accounting for the causal structure of real-world data is rarely trivial since the causal mechanisms that gave rise to the data are typically unknown. To address this problem, we develop a feature selection method that considers each feature's value for HTE estimation and learns the relevant parts of the causal structure from data. We provide strong empirical evidence that our method improves existing data-driven HTE estimation methods under arbitrary underlying causal structures. Our results on synthetic, semi-synthetic, and real-world datasets show that our feature selection algorithm leads to lower HTE estimation error.
\end{abstract}



\maketitle

\section{Introduction} \label{sec:intro}

Estimating heterogeneous treatment effects (HTEs) is essential in many fields.
HTE estimation aims to find subpopulations whose causal effects differ from the effects of the population as a whole.
For example, if the treatment is a drug, some individuals may have adverse reactions, and some individuals may benefit from treatment~\cite{shalit-icml17}. 
Similarly, public policy may affect different sociodemographic groups differently~\cite{grimmer-pa17}.
HTE analysis allows the discovery of these different subgroups of the population.

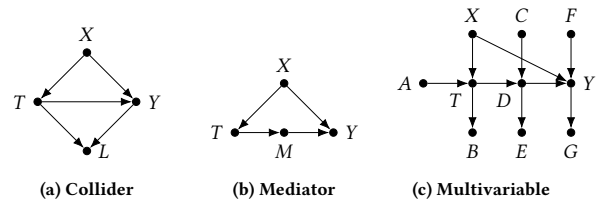
\begin{figure}
    \centering
    \newcommand{\scmsize}{0.3}
    \resizebox*{\scmsize\columnwidth}{!}{
        \begin{subfigure}[b]{0.325\columnwidth}
            \centering
            \begin{tikzpicture}
                \node (t) at (-0.75, 0) [label=left:$T$,point];
                \node (x) at (0.0, 0.75) [label=above:\(X\),point];
                \node(y) at (0.75, 0.0) [label=right:$Y$,point];
                \node(c) at (0.0, -0.75) [label=right:$L$,point];

                \path (x) edge (t);
                \path (x) edge (y);
                \path (t) edge (y);
                \path (t) edge (c);
                \path (y) edge (c);
            \end{tikzpicture}
            \caption{Collider}\label{fig:collider_scm}
        \end{subfigure}
    }
    \resizebox*{\scmsize\columnwidth}{!}{
        \begin{subfigure}[b]{0.325\columnwidth}
            \centering
            \begin{tikzpicture}
                \node (t) at (-0.75, 0) [label=left:$T$,point];
                \node (m) at (0.0, 0) [label=below:$M$,point];
                \node (x) at (0.0, 0.75) [label=above:\(X\),point];
                \node(y) at (0.75, 0.0) [label=right:$Y$,point];

                \path (x) edge (t);
                \path (x) edge (y);
                \path (t) edge (m);
                \path (m) edge (y);
            \end{tikzpicture}
            \caption{Mediator}\label{fig:mediator_scm}
        \end{subfigure}
    }
    \resizebox*{\scmsize\columnwidth}{!}{
        \begin{subfigure}[b]{0.325\columnwidth}
            \centering
            \begin{tikzpicture}
                \node (a) at (-1.5, 0) [label=left:$A$,point];
                \node (t) at (-0.75, 0) [label={below left}:$T$,point];
                \node (b) at (-0.75, -0.75) [label={below}:$B$,point];
                \node (x) at (-0.75, 0.75) [label={above}:$X$,point];
                
                \node (c) at (0.0, 0.75) [label={above}:$C$,point];
                \node (d) at (0.0, 0) [label={below left}:$D$,point];
                \node (e) at (0.0, -0.75) [label={below}:$E$,point];
        
                \node (f) at (0.75, 0.75) [label={above}:$F$,point];
                \node (y) at (0.75, 0.0) [label={right}:$Y$,point];
                \node (g) at (0.75, -0.75) [label={below}:$G$,point];
        
                \path (a) edge (t);
                \path (x) edge (t);
                \path (t) edge (b);
                
                \path (x) edge (y);
                \path(t)edge(d);
                \path(c)edge(d);
                \path(d)edge(e);
                \path (d) edge (y);
        
                \path (f) edge (y);
                \path (y) edge (g);
            \end{tikzpicture}
            \caption{Multivariable}\label{fig:marloes_scm}
        \end{subfigure}
    }
    \caption{Causal models where data-driven HTE estimation methods may perform poorly. 
    Some variables (e.g., $L$, $M$, and $E$) are not valid adjustment variables for estimating the effect of $T$ on $Y$.
    }\label{fig:fail_examples}
\end{figure}

The accumulation of large amounts of data in many domains has stimulated the increased development of algorithms tailored for automated \emph{data-driven} HTE estimation, relying on supervised machine learning~\cite{athey-pnas16,johansson-icml16,grimmer-pa17,shalit-icml17,kennedy-arxiv20,kunzel-pnas19,tran-aaai19,athey-annals19}. 
The goal of data-driven HTE estimators is to estimate the conditional average treatment effect using the features available in the data. 

While it is appropriate to use all possible features for general supervised learning problems, unbiased estimation of \emph{causal effects} requires knowledge of the causal mechanisms underlying the data and selecting an appropriate subset of variables from the data~\cite{pearl-book09}.
One way to select this subset is by modeling the underlying causal mechanisms through structural causal models (SCMs)~\cite{pearl-book09} and selecting an \emph{adjustment set} --- a set of variables that are necessary for unbiased estimation of a causal effect of interest~\cite{pearl-book09,perkovic-uai15,henckel-arxiv19}. 
However, the causal model structure is rarely known for real-world data, and data-driven methods do not take it into consideration. 
Our paper addresses this problem and improves data-driven HTE estimation methods through causal feature selection under such structure uncertainty.

A common assumption used by data-driven HTE estimators is the assumption of \emph{strong ignorability} which states that the treatment assignment is independent of the potential outcomes given the variables in the data~\cite{rosenbaum-bio83,shalit-icml17}.
This untestable assumption implies that all potential confounders, variables that can influence both the treatment and the outcome, are observed in the data.
In turn, accounting for them would remove spurious associations and allow for causal effect identification.
However, hidden confounding is not the only problem leading to biased HTE estimation. 
Even if all potential confounders are observed in the data, not all variables should be used for adjustment.
Another important and sometimes implicit assumption is that all variables in the data are \emph{pre-treatment} variables, i.e., the variables that were measured before and thus were unaffected by the treatment.
When the temporal order of observations is unknown, many practitioners would simply use all variables in the data~\cite{montgomery-ajps18}.
To illustrate the problem of using all variables in data-driven methods, consider the SCMs in Figure~\ref{fig:fail_examples} where $X$ is a confounding variable. 
According to SCM theory, we need to adjust for \( X \) when estimating causal effects, but should not include variables that will bias effect estimation, such as descendants of the treatment or instrumental variables (e.g., \( L, M, A, B, D, E, G \)).

To address these deficiencies of data-driven methods, we define the problem of causal feature selection for accurate data-driven HTE estimation under structure uncertainty.
One vanilla solution to this problem is to first learn the causal structure from the data~\cite{henckel-arxiv19,spirtes-book00,ogarrio-pgm16} and then use an existing identification algorithm to find the adjustment set~\cite{shpitser-uai06,van-uai14,perkovic-uai15,henckel-arxiv19}.
However, structure learning is computationally expensive when the space of feature is large, and it may not work well when there is uncertainty in the learned structure~\cite{heinze-stat18,chickering-jmlr04,vowels-arxiv21}.
Instead, we develop a feature selection method that reduces this complexity by considering the value of each feature for improving HTE estimation and learning only the relevant parts of the causal structure from data. 
We provide strong empirical evidence that our method improves existing data-driven HTE estimation methods under arbitrary underlying causal structures.
We evaluate the benefits of our feature selection algorithm in reducing HTE estimation error through synthetic, semi-synthetic, and real-world dataset experiments.

\section{Related Work}

We provide a brief overview of related work on estimating heterogeneous treatment effects and structure learning.

\textbf{Data-driven methods for HTE estimation.}
Many methods have been developed for estimating heterogeneous treatment effects (HTEs) from data.
Several methods rely on recursive partitioning using tree-based methods, which naturally partition the population into subgroups that may contain heterogeneous effects~\cite{athey-pnas16,athey-annals19,tran-aaai19,su-jmlr09,zeileis-jcgs08}.
Besides tree-based methods, methods have been developed that use machine learning to predict causal effects, such as meta-learners~\cite{kunzel-pnas19,kennedy-arxiv20} and neural networks~\cite{johansson-icml16,shalit-icml17,schwab-perfect18}.
However, these methods do not consider the structure of the underlying causal mechanisms that generate the data, introducing bias in estimation.

\textbf{Causal structure learning.}
Many causal structure learning algorithms have been developed for recovering SCMs from data~\cite{heinze-stat18,vowels-arxiv21}.
Heinze-Deml et al.\ provide a recent survey on causal structure learning algorithms and evaluate their performance in various settings~\cite{heinze-stat18}.
However, learning the full causal structure from data can be computationally expensive for high-dimensional datasets~\cite{chickering-jmlr04,vowels-arxiv21}.
Related to causal structure learning is local structure learning~\cite{aliferis-jmlr10}.
Aliferis et al.\ presented an algorithmic framework for learning local causal structure around target variables, the parent-children set (\( PC \)) and the Markov Blanket (\( MB \)).
Yu et al.\ survey and evaluate several local causal structure learning methods that learn both \( PC \) and \( MB \) of target variables~\cite{yu-arxiv19}.
They study local structure learning under the context of causality-based feature selection with the goal of improving supervised learning. 
In contrast, we study the problem of causal feature selection.
Local structure learning has been studied in the context of average treatment effect (ATE) estimation, but not HTE estimation~\cite{cheng-arxiv20}.
This work also does not consider excluding mediators or descendants, which we do in our work.

\section{Background}\label{sec:background}

We provide an introduction to heterogeneous treatment effects under the potential outcomes framework~\cite{rubin-74} and structural causal models~\cite{pearl-book09}.

\subsection{Heterogeneous treatment effects}

Let $(\features_i, Y_i(0), Y_i(1), T_i) \sim \mathcal{P}$ be a distribution with $N$ units which are independently and identically distributed (i.i.d.).
For any unit $i$, define $\features_i \in \mathbb{R}^d$ be a $d$-dimensional feature vector, 
$T_i \in \{0, 1\}$ to be the treatment assignment indicator, $Y_i(0)$ to be the potential outcome when $i$ is not treated, and $Y_i(1)$ to be the potential outcome when $i$ is treated.
The individual treatment effect (ITE) is the difference in potential outcomes: \( \tau_i = Y_i(1) - Y_i(0) \).
However, both potential outcomes $Y_i(1)$ and  $Y_i(0)$ cannot be observed at the same time and observed data $\mathcal{D} = (\features_i, Y_i, T_i)$ contains only $Y_i = Y_i(T_i)$. This is known as the fundamental problem of causal inference.
Since $\tau_i$ can never be measured directly, its estimate is expressed through the conditional average treatment effect (CATE):
\begin{equation}
    \label{eq:cate}
    \hat{\tau}(\mathbf{x}) = E[Y(1) - Y(0) \mid \features_i].%
\end{equation}
The average treatment effect (ATE) of the population is: $\text{ATE} = E[Y(1) - Y(0)]$. If CATE and ATE are not the same, then $\features_i$ induce heterogeneous treatment effects (HTE).

Given an observed dataset $\mathcal{D}$, the goal of an \emph{HTE estimator} is to estimate CATE, $\hat{\tau}$, using the available features $\features$ in the data.
HTE estimators estimate CATE in a variety of ways.
Some HTE estimators rely on recursive partitioning using tree-based methods~\cite{tran-aaai19,athey-pnas16,athey-annals19}.
Other HTE estimators use predictions from machine learning methods to train a causal effect estimator~\cite{kunzel-pnas19,kennedy-arxiv20} or to predict counterfactual outcomes for estimation of effect~\cite{shalit-icml17,johansson-icml16}.
A common assumption used by data-driven HTE estimators is the assumption of \emph{strong ignorability} which states that the treatment assignment is independent of the potential outcomes given a set of variables $\mathbf{X}$ ~\cite{rosenbaum-bio83,shalit-icml17}: $Y_1, Y_0 \perp T \mid \mathbf{X}$, and $0 < p(T=1 \mid \mathbf{X}) < 1$ for all $\features = \mathbf{x}$.
One of the implications of this assumption is that there is no \emph{hidden confounding}.
However, hidden confounding is not the only problem that can lead to biased HTE estimation. In practice, it is often assumed that all observed variables in the data are the set of variables $\mathbf{X}$ that meets strong ignorability.
This is problematic because some of the variables may make $T$ and $Y$ dependent or introduce new spurious dependencies~\cite{pearl-book09}. 
Some variables, such as mediators, should not be included in the estimation. Moreover, when the temporal order of variables is unknown, practitioners may consider all features in the data as pre-treatment variables~\cite{montgomery-ajps18}.
Selecting the right variables is the goal of our work.

\subsection{Structural Causal Models}

A structural causal model (SCM), $M$, consists of two sets of variables, $\mathbf{U}$ and $\mathbf{V}$, and a set of structural equations, $\mathbf{F}$, describing how values are assigned to each \emph{endogenous} variable $V_i \in \mathbf{V}$ based on the values of $\mathbf{v}$ and $\mathbf{u}$~\cite{pearl-book09}: \( v_i = f_i(\mathbf{v}, \mathbf{u}) \).
A causal graph \( G \) is a directed acyclic graph that captures the causal relationships among the variables.
The variables in $\mathbf{U}$ are considered \emph{exogenous}.
It is assumed that every endogenous variable has a parent in $\mathbf{U}$.
The causal graph may omit representing $\mathbf{U}$ explicitly.
We denote the parents, children, ancestors, and descendants of $X$ in the graph as: \(pa(X), ch(X), an(X), de(X)\).
Every node is a descendant and ancestor of itself.

When the SCM underlying the dataset is known, its graph G supports the use of graphical criteria for variable selection for unbiased causal effect estimation, known as adjustment set selection~\cite{shpitser-uai06,shpitser-uai10,van-uai14,perkovic-uai15,henckel-arxiv19}.
One such criterion is the backdoor criterion~\cite{pearl-book09}, which aims to find a set that blocks the paths from treatment to the outcome, starting with an arrow into treatment.
For example, in Figure~\ref{fig:collider_scm}, there is one backdoor path, namely, $ T \leftarrow X \rightarrow Y $, and so the set $ \{ X \} $ sufficiently satisfies the backdoor criterion.
Several adjustment criteria have been developed, such as the adjustment criterion~\cite{shpitser-uai06,van-uai14}, generalized back-door criterion~\cite{maathuis-annals15}, generalized adjustment criterion~\cite{perkovic-uai15}, and optimal adjustment set (O-set)~\cite{henckel-arxiv19}.
These adjustment sets defined by the various adjustment criteria can be found through the ID algorithm~\cite{shpitser-jmlr08}, the framework by Van der Zander et al.~\cite{van-uai14} or the pruning procedure by Henckel et al.~\cite{henckel-arxiv19}.
Once an adjustment set is found, they can be used with an HTE estimator to estimate heterogeneous effects.

When the SCM for a dataset is unknown, and therefore there is \emph{structure uncertainty} of the causal mechanism, the adjustment criteria cannot be applied.
An alternative approach is to use causal structure learning~\cite{heinze-stat18} to learn a causal graph from the data and then apply an adjustment criterion for feature selection.
However, learning the full structure is computationally expensive for high-dimensional datasets~\cite{chickering-jmlr04,heinze-stat18,vowels-arxiv21}.
Instead, we propose reducing the feature space by assessing the fit of each variable for HTE estimation and learning the local structure, which we describe in Section~\ref{sec:cfs}.

\subsubsection{HTEs in SCMs}

In SCMs, heterogeneous effects are represented by an \emph{interaction} or \emph{effect modification} through ancestors of $Y$~\cite{pearl-book09}.
For example, suppose $Y$ is generated by the following structural equation corresponding to the SCM in Figure~\ref{fig:marloes_scm}:
\begin{equation}
    Y = \alpha \cdot T + \beta \cdot X + \gamma \cdot T \cdot f(X) + U_Y.
\end{equation}
Here, there is a fixed effect \( \alpha \) of the treatment \( T \) and a heterogeneous effect \( \gamma \) due to the interaction between \( T \) and \( X \).
Interaction can occur with any parent of \( Y \) (e.g., \( X, F \) in Figure~\ref{fig:marloes_scm}) or any parent of a mediator (e.g., \( C \) in Figure~\ref{fig:marloes_scm})~\cite{vanderweele-epid07}.

\subsubsection{C-specific effects}

\new{
Shpitser and Pearl provide a nonparametric estimand for HTEs with respect to some feature assignment $C=c$, called $c$-specific effects~\cite{shpitser-uai06,pearl-smr17}.
}
One issue with estimating $c$-specific effects is that it requires knowledge of the SCM and which variables, C, exhibit heterogeneous effects.
When the SCM is known, the O-set~\cite{henckel-arxiv19} can be used on the learned structure since it implicitly includes all potential heterogeneity-inducing variables, while some identification algorithms may miss them. When the SCM is unknown, the identifiability of \( c \)-specific effects cannot be established. 
Moreover, when the set of heterogeneity-inducing variables $\mathbf{C}$ is unknown, then all combinations of features and feature values would need to be checked for heterogeneity in effect, which is exponential in the number of valid features. 
Circumventing this exponential search is one of the main reasons why data-driven methods are popular in practice.

\section{Problem Setup}

Our goal in this work is to augment data-driven methods to accurately predict heterogeneous effects when both the SCM and the variables that explain heterogeneity are unknown.
More specifically, the goal is to find a subset $\mathbf{X^{(\ell)}}$, to which we refer as the \emph{\subsetname{}}, that improves data-driven estimation.
In general, the goal in HTE estimation is to obtain accurate estimation of CATE:
\begin{problem}\label{prob:hte_problem}
    (Heterogeneous treatment effect (HTE) estimation)
    Given a dataset $\mathcal{D}=(\features_i, Y_i, T_i)$ of N instances with some true ITE $\mathbf{\tau}$, estimate CATE $\hat{\tau}(\mathbf{X})$ such that the mean squared error is minimized:
    \begin{equation}\label{eq:cate_objective}
         \frac{1}{N} \sum_{i=1}^N  (\tau_i - \hat{\tau}(\mathbf{X}_i)) ^ 2.
    \end{equation}
\end{problem}
Since the true ITE is unknown in real-world scenarios, several heuristic HTE evaluation metrics have been developed for evaluating the performance of HTE estimators~\cite{schuler-model18,schwab-perfect18,saito-cv19}.
Instead of minimizing the true estimator error, we can instead minimize one of these metrics, such as the \( \tau \)-risk~\cite{schuler-model18,nie-rlearner17}.

A \emph{\subsetname{}} is the feature set that minimizes the error in HTE estimation.
\begin{problem}\label{prob:causal_feature_selection}
    (Causal feature selection (CFS) for HTE estimation) Given a dataset $\mathcal{D}=(\features_i, Y_i, T_i)$ and a data-driven estimator $\hat{\tau}(\mathbf{X})$, find a \emph{\subsetname{}} $\features_i^{(\ell)} \subseteq \features_i$ such that $\hat{\tau}(\mathbf{X}^{(\ell)})$ results in the lowest MSE:
    \begin{gather}
        \argmin_{\mathbf{X^{(\ell)}}} \frac{1}{N} \sum_{i=1}^{N} {\Big( \tau_i - \hat{\tau}\big( \mathbf{X}^{(\ell)}_i \big) \Big)}^2.
    \end{gather}
\end{problem}
\noindent A data-driven estimator typically estimates $\hat{\tau}(\mathbf{X})$ using all available data. 
Our goal in this work is to select a \subsetname{} $\mathbf{X}^{(\ell)}$ such that the error of the estimator is minimized.

In this work, we assume that it is unknown a priori whether all observed features in the data are pre-treatment variables and whether all of them meet the strong ignorability assumption.
Instead, we use the following more general assumptions:
\begin{enumerate}
    \item Causal sufficiency: there are no latent variables.
    \item Causal faithfulness: if $X_A, X_B$ are conditionally independent given $\mathbf{X_Z}$, then $A,B$ are d-separated in the causal graph by $\mathbf{Z}$.
    \item Acyclicity: the underlying causal graphs are directed acyclic graphs (DAGs).
    \item There is no selection bias.
\end{enumerate}
Understanding how our methods need to be adjusted when these assumptions do not hold is left for future work.

\begin{figure*}[!ht]
    \includegraphics[width=\linewidth]{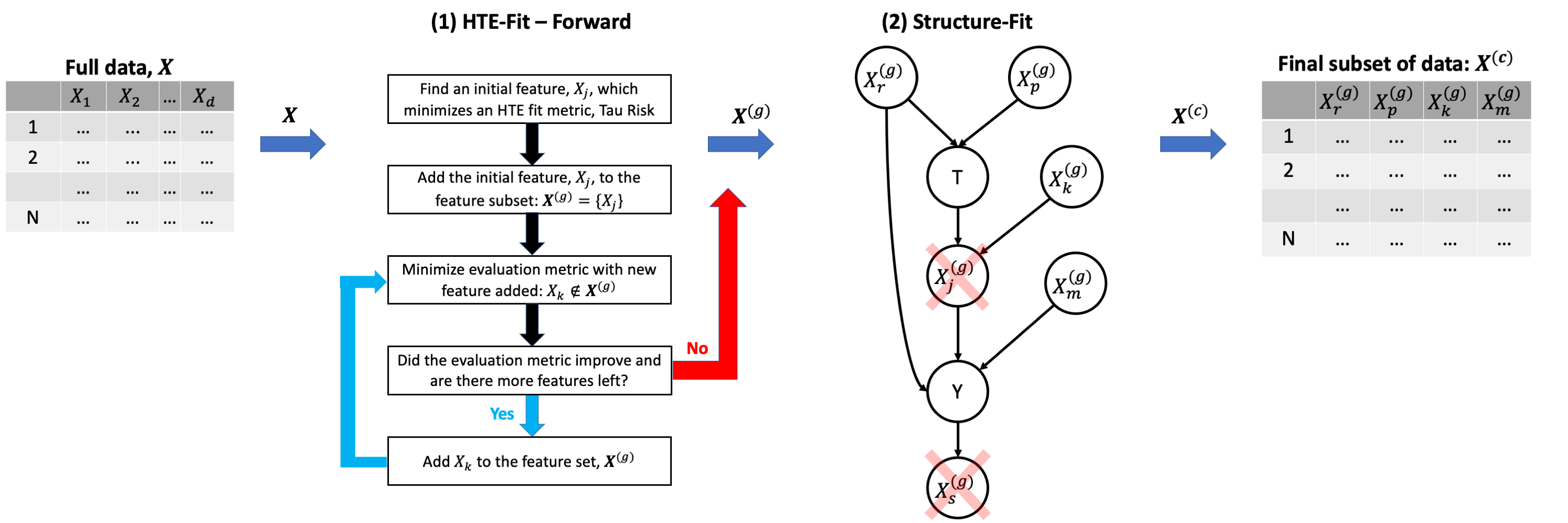}
    \vspace{-2em}
    \caption{Our causal feature selection method \combined{}. 
    Given the full set of features $\mathbf{X}$, the first step in \combined{} is to select a subset of features that minimizes an HTE fit metric (e.g., Tau risk).
    Then, given the subset of features found, $\mathbf{X}^{(g)}$, a structure is learned on the path from $T$ to $Y$.
    The final feature set is found by removing mediators and descendants of treatment (crossed out in red), which results in the final feature subset of $\mathbf{X}^{(c)} \subseteq \mathbf{X}$.
    }\label{fig:glcfs_fig}
\end{figure*}

\section{Causal Feature Selection for HTE Estimation}\label{sec:cfs}

When the SCM that generated the data is unknown, a vanilla solution for finding an adjustment set for HTE estimation is to learn the causal structure and then find the optimal adjustment set (O-set)~\cite{henckel-arxiv19}.
However, structure learning is computationally expensive in high-dimensional data, both in theory~\cite{chickering-jmlr04} and in practice~\cite{heinze-stat18,vowels-arxiv21}.
Instead of doing full structure learning, we propose \Combined{}, which relies on two ideas to make feature selection more practical.
First, instead of considering all features, it assesses the value of each feature in reducing the HTE estimation error and selects only the features that contribute to error reduction. 
Second, instead of learning the full structure, it focuses on learning the parts of the structure that give sufficient information about selecting a valid adjustment set.
Figure~\ref{fig:glcfs_fig} shows a high-level overview of \combined{}, and we present its main components, \Globalmethodname{} and \Localmethodname{}, next.

\subsection{\Globalmethodnamenoparen{}}

The first component of our framework, \Globalmethodname{}, proposes the use of \emph{heuristic HTE evaluation} metrics to sequentially select features that are likely to reduce the HTE estimation error.
This is shown as the first step in Figure~\ref{fig:glcfs_fig}.
It has two variants: forward (\globalforward{}) and backward (\globalbackward{}).

Since the ground truth of HTE effects in unknown, several heuristic HTE evaluation metrics have been developed to approximate the errors of HTE estimators, including $\tau$-risk~\cite{schuler-model18}, Nearest Neighbor Precision in Estimating Heterogeneous Effects (NN-PEHE)~\cite{schwab-perfect18}, Plug-in $\tau$~\cite{saito-cv19}, and Counterfactual Cross-validation (CFCV)~\cite{saito-cv19}. 
The $\tau$-risk takes as input the estimator $\hat{\tau}$ and approximates the objective defined in~\eqref{eq:cate_objective}~\cite{schuler-model18}.
The second metric, NN-PEHE, uses nearest neighbor matching for imputing treatment effects for training~\cite{schwab-perfect18}.
The plug-in $\tau$ uses an arbitrary estimator for imputing effects for computing error.
The fourth metric, CFCV, uses a doubly robust formulation with the plug-in $\tau$ for imputing effects for evaluation~\cite{saito-cv19}.
These metrics have been shown to correlate with HTE estimator performance~\cite{schuler-model18,saito-cv19}. 
Here, we propose to use these metrics as causal feature selection criteria, which, to the best of our knowledge, has not been studied in previous research.

\subsubsection{\Globalforward{}}

\globalforward{} starts with a feature with the best (lowest) error metric, then iteratively adds features that improve the metric.
If there are no more features or improvements, the selection stops.
We will use $\tau$-risk as an example metric to make this procedure more concrete. $\tau$-risk has been shown to lead to a consistent high-performing model~\cite{schuler-model18} and to have quasi-oracle error bounds on the CATE~\cite{nie-rlearner17}.
The $\tau$-risk, denoted as $R$, takes as input the estimator $\hat{\tau}$ and approximates the objective defined in~\eqref{eq:cate_objective}~\cite{schuler-model18}:
\begin{equation}\label{eq:tau_risk}
    \tau\text{-risk}(\hat{\tau}) = R(\hat{\tau}) = \frac{1}{N} \sum_{i=1}^N \Big( 
        (Y_i - \hat{m}(\mathbf{X}_i)) - (T_i - \hat{p}(\mathbf{X}_i)) \hat{{\tau}}(\mathbf{X}_i)
    \Big)^2.
\end{equation}
Here, \( \hat{m}(\mathbf{X}_i) \) is an outcome estimate for unit $i$ (i.e., \( E[Y_i | \mathbf{X}_i] \)),  and \( \hat{p}(\mathbf{X}_i) \) is an estimate of the propensity score, (i.e., \( E[T_i | \mathbf{X}_i] \)).
Both $\hat{m}$ and $\hat{p}$ are learned through supervised techniques and optimized for predictive accuracy.

Suppose the feature vector of individuals has $d$ dimensions, $\features = (X_1, \dots, X_d)$.
Given the $\tau$-risk, we can compute the feature fit for one feature, say $X_j \in \features$, by estimating $\hat{\tau}(\{X_j\})$ and inputting the estimator into the $\tau$-risk.
For \globalforward{}, the first feature found is the feature that minimizes $R(\hat{\tau}(\{X_j\})$, and we define the initial set of found features as $\features^{(g)} = \{X_j\}$.
Now that the first feature is found, we need to add another to $\features_{(g)}$.
For each feature $X_k \in \features, X_k \notin \features^{(g)}$, we estimate a new $\hat{\tau}(\features^{(g)} \cup \{X_k\})$ which is the current set of found features $\features^{(g)}$ combined with a new feature.
The next feature added to $\features^{(g)}$ is the feature that minimizes $R(\hat{\tau}(\features^{(g)} \cup \{X_k\})$.
This process continues until there is no feature that, when combined with $\features^{(g)}$, minimizes the $\tau$-risk or when there are no more features to add.

\subsubsection{\Globalbackward{}}

Backward feature selection (\globalbackward{}) works similar to forward feature selection.
The difference is that it starts with all features, $\features^{(g)} = \features$, and it
iteratively removes features until either the error metric does not improve or there is only one feature left for estimation.
Using $\tau$-risk, it would first compute $R(\hat{\tau}(\features))$ to get an initial score on the entire feature set.
For each $X_j \in \features$, it would compute $R(\hat{\tau}(\features \setminus \{ X_j \})$, and remove the feature that results in the lowest $\tau$-risk when removed.

\subsection{\Localmethodnamenoparen{}}

Now that we have selected the features that can contribute to HTE estimation error reduction, the next step is to select only the features which are a valid adjustment set. The second part of our framework, 
\localmethodname{}, iteratively applies local structure learning~\cite{aliferis-jmlr10,yu-arxiv19} to find a path from the treatment, $T$, to the outcome $Y$, in order to remove mediators and descendants of treatment.
Step 2 of Figure~\ref{fig:glcfs_fig} represents this part, \localmethodname{}, where local structures from $T$ to $Y$ are learned iteratively, and connected. A local structure is a subgraph of the causal graph centered on a target variable.
For example, in Figure~\ref{fig:glcfs_fig}, the local structure of $T$ are the edges and variables connected to $T$ (e.g., $X_r^{(g)}, X_p^{(g)}, X_j^{(g)}$).
Since $Y$ is the outcome variable and we are only interested in paths from $T$ to $Y$, \localmethodname{} terminates and the final feature set is found by removing mediators and descendants (crossed out in red).

This approach has two main advantages compared to full causal structure learning.
First, it only considers local structure around causal nodes and does not need to consider all variables if they are not connected to the causal path from \( T \) to \( Y \).
Second, it aims to only select variables which do not bias causal effect estimation and allow the HTE estimator to use features as appropriate for estimation.

\subsubsection{Discovering local structure around target variables}
Multiple structures can encode the same conditional independences in the data, making the task of finding the parent sets of $T$ and $Y$, $pa(T), pa(Y)$, nontrivial. 
For example, $X_1 \rightarrow T \rightarrow X_2, X_1 \leftarrow T \leftarrow X_2$, and $X_1 \leftarrow T \rightarrow X_2$, all encode that $X_1 \perp X_2 | T$ and in one of them $X_1$ is a parent of $T$ whereas in the other two $X_1$ is a child.

We use two different ways to find the local structure.
The first way relies on finding the parent-children ($PC$) set around our target variables, $T$ and $Y$~\cite{aliferis-jmlr10}, collider discovery, and edge orientation methods to find causal directions~\cite{daniusis-uai12,fonollosa-book19,blobaum-aistats18}.
Several $PC$ discovery algorithms have been developed~\cite{buhlmann-bio10,pena2005scalable,aliferis-jmlr10,yu-arxiv19} and have mainly been used for improving classification methods~\cite{yu-arxiv19}.
To distinguish between parents and children, we first discover the parents of each target variable through collider discovery.
If two variables are parents of the target variable (e.g., \( X \) and \( A \) are parents of \( T \)), then they are dependent if conditioning on the child and can be discovered through conditional independence tests.
However, if there is zero or one parent, we will not leverage collider discovery.
Instead, we orient the remaining edges using an edge orientation algorithm~\cite{daniusis-uai12,fonollosa-book19,blobaum-aistats18} and then check if there is a parent in the set.
Using these algorithms, we can check for any variable in $PC$ whether an arrow is pointing in or away from the target variable (i.e., $T$ and $Y$) and thus identify $pa(T), pa(Y), ch(T)$, and $ch(Y)$.

The second way to find local structures is to utilize more recent local structure learning algorithms, such as PCD-by-PCD~\cite{yin-causation08}, MB-by-MB~\cite{wang-csda14}, and Causal Markov Blanket~\cite{gao-neurips15}
These algorithms discover the $PC$ or Markov Blanket ($MB$) while partially orienting edges.
We can find a partially oriented local structure and orient edges using edge orientation methods.

\subsubsection{Learning partial structures}

Using the process to discover local structure around target variables, we propose an iterative algorithm to remove post-treatment variables from \globalmethodname{}, which we call \Localmethodname{}.
We apply local structure learning on the treatment variable, $T$, and then iteratively apply local structure learning on children of $T$ in a breadth-first-search manner until we reach the outcome variable $Y$.
Then, local structures are put together to form a subgraph of the causal graph over the variables selected by \globalmethodname{}.
After learning the local structure, we can apply an identification algorithm to find an adjustment set, such as the O-set~\cite{henckel-arxiv19}.
Another option is to only remove all descendants of treatment.

We demonstrate \localmethodname{} using Figure~\ref{fig:marloes_scm}.
We first find $PC(T)=\{A, X, B, D\}$ and $ch(T)=\{B, D\}$.
We proceed to iteratively find the $PC$ on each child, $B, D$.
Since $B$ has no children, we move to $D$ and find $PC(D)=\{T, C, E, Y\}$ and $ch(D)=\{E, Y\}$.
Finally, $E$ has no children, and we have reached the outcome variable, $Y$ and find $PC(Y)=\{X, C, F, G\}$.
Using the local structures, we find the causal graph shown in Figure~\ref{fig:marloes_scm}.
In practice, we only use the output of \globalmethodname{} in \localmethodname{}, so not all variable will be used.

While \globalmethodname{} and \localmethodname{} are both important parts of the framework, each one can be used independently for causal feature selection as well.
However, each component has limitations, which can be addressed by combining them into one framework.
One limitation of \globalmethodname{} is that it is based on a heuristic evaluation metric for HTE estimation.
While the evaluation metric is correlated with the true performance~\cite{schuler-model18}, these metrics may still include non-valid adjustment variables.
This issue can be addressed by discovering mediators and descendants of treatment and excluding them from the final estimation.
One limitation of using \localmethodname{} is that variables that are potentially irrelevant for HTE estimation are considered in the local structure learning, which can lead to noisy independence tests and discovered structures.
This is addressed by finding the most important subset of features for HTE estimation.
\combined{} takes advantages of \globalmethodname{} and \localmethodname{} to address limitations of both methods.
Pseudocode of \globalmethodname{} and \localmethodname{} is available in the Appendix in Sections~\ref{sec:hte_fit_desc}~and~\ref{sec:structure_fit_desc}, respectively, and a discussion of complexity is available in Section~\ref{sec:complexity}.

\section{Experiments}

To evaluate whether and how much feature selection improves existing data-driven HTE algorithms, we generate synthetic datasets which contain ground-truth causal effects.
We also use a semi-synthetic dataset and two real-world datasets.

\subsection{Experimental setup}

In our experiments, we make a distinction between four classes of estimation methods: \emph{base learners}, \emph{baseline feature selection}, \emph{our feature selection}, and \emph{oracle}.
A base learner is an HTE estimator without feature selection, and it uses all features for estimation.
We consider several prominent HTE estimators as base learners: 
Causal Tree Learn~\cite{tran-aaai19}, Causal Forest~\cite{athey-annals19},
Meta-learners (S, T, XLearners)~\cite{kunzel-pnas19}, Doubly Robust learner (DRLearner)~\cite{kennedy-arxiv20}, Balancing Neural Network (BNN)~\cite{johansson-icml16}, and TARNet~\cite{shalit-icml17}.

The baseline feature selection methods use local structure learning~\cite{yu-arxiv19} and causal structure learning~\cite{heinze-stat18,henckel-arxiv19} to select an adjustment set.
The first baseline, \emph{PC Simple (Parents)}, uses a local structure learning method on the treatment to find parents. 
PC Simple has been used for selecting variables for classification~\cite{yu-arxiv19} but not for HTE estimation.
We employ PC Simple to learn the local structure on the treatment variable and select the treatment parents as an adjustment set.
Since our setting only considers one treatment variable, the parents of treatment are a valid adjustment set~\cite{williamson-resp14}.
Two other baselines we consider are based on causal structure learning~\cite{heinze-stat18,henckel-arxiv19}, which we refer to as \Fullstructure{}.
We use GFCI~\cite{ogarrio-pgm16} to learn a causal structure and apply two adjustment criteria: optimal adjustment set (O-set) and removing post-treatment variables only (Valid).
Then a base learner uses the adjustment set for estimation (e.g., TLearner).

For our feature selection methods, we consider our framework \combined{} and each component, \globalmethodname{} and \localmethodname{}, separately.
For \globalmethodname{}, we use both variants: \globalforward{} and \globalbackward{}.
We use four HTE fit metrics: $\tau$-risk~\cite{schuler-model18}, Nearest Neighbor Precision in Estimating Heterogeneous Effects (NN-PEHE)~\cite{schwab-perfect18}, plug-in $\tau$~\cite{saito-cv19}, and Counterfactual Cross-validation (CFCV)~\cite{saito-cv19}.
We report results using \globalforward{} and $\tau$-risk since this combination performed the best in our experiments.

For \localmethodname{} we used PC-Simple~\cite{buhlmann-bio10}, GetPC~\cite{pena2005scalable}, and HITON-PC~\cite{aliferis-jmlr10}, CMB~\cite{gao-neurips15}, and PCD-by-PCD~\cite{yin-causation08}.
For pairwise edge orientation, we use Conditional Distribution Similarity Statistic (CDS)~\cite{fonollosa-book19}, Information Geometric Causal Inference (IGCI)~\cite{daniusis-uai12}, and Regression Error based Causal Inference (RECI)~\cite{blobaum-aistats18}.
In order to find an adjustment set, we only remove mediators and descendants, and allow data-driven HTE estimators to decide which variables to use in estimation.
We report results using PC Simple with RECI (PC, RECI) and CMB since these two combinations represent different variants of \localmethodname{} and perform the best in our experiments.

For \combined{}, we report results using \globalforward{} with $\tau$-risk and \localmethodname{} with CMB, since this combination performed the best in experiments.
We compare against using Oracle (Known Structure), which uses the true optimal adjustment set.

\subsection{Datasets}

We use three types of datasets.
The first dataset is a set of synthetic datasets, in which random SCMs are generated, and data is generated following each SCM.
The second dataset is based on the Infant Health and Development Program (IHDP) dataset~\cite{hill-jcgs11}.
The third type are real-world datasets, League of Legends and Cannabis.

\subsubsection{Synthetic dataset}

We generate SCMs by following a procedure similar to one for evaluating causal structure learning algorithms~\cite{heinze-stat18}.
SCMs and their corresponding datasets differ in the following characteristics: the number of variables \( d \); the probability of edge creation \( p_e \); the variance in the noise \( \sigma \); the strength of the noise term in non-source nodes \( \rho \).
We add parameters to control confounding \( \gamma \), mediation \( m \), and heterogeneity in treatment effects \( p_h, m_p \).
We initialize a set of parameters as follows:
\begin{enumerate}
    \item Number of variables: \( d \in [10, 20, 30] \)
    \item Edge probability parameter: \( p_e \in [0.1, 0.3, 0.5] \)
    \item Noise variance: \( \sigma \in [0.2, 0.4, 0.6] \)
    \item Magnitude in noise: \( \rho \in [0.1, 0.5, 0.9] \)
    \item Confounder: \( \gamma \in [\text{True, False}] \)
    \item Mediator chain length: \( m \in [0, 1, 2] \)
    \item Number of HTE inducing parents: \( p_h \in [0, 1, 2] \)
    \item HTE from mediating parents: \( m_p \in [\text{True, False}] \)
\end{enumerate}

For each parameter value, we randomly sample from all other parameter values \( 500 \) times to generate SCMs.
For each of the $11,500$ SCMs, we create one dataset of size \( 10,000 \).
Root node variables are generated from a normal distribution, and non-root nodes are generated linearly from parents.
The detailed dataset generation description is available in the Appendix in Section~\ref{sec:data_gen_desc}, and the link for the code is available in Section~\ref{sec:code}.

\subsubsection{Semi-synthetic dataset: IHDP}
The covariates in the Infant Health and Development Program (IHDP) dataset come from a randomized experiment studying the effects of specialist home visits on future cognitive test scores~\cite{hill-jcgs11}.
Outcomes can be simulated using the covariates.
We simulate 1,000 instances of simulated outcomes and average all results\footnote{Outcomes are generated using the JustCause package:\\https://github.com/inovex/justcause}.

\subsubsection{Real-world datasets}

\kdd{
We use two real-world datasets: \emph{League of Legends} and \emph{Cannabis}.
League of Legends (LoL) is a popular multiplayer online battle arena (MOBA) game.
Riot Games, the developer of LoL, regularly creates software patches to the game that update various parts of the game, affecting balance and how players decide to play the game.
We use a dataset consisting of player, match, and patch information over time~\cite{he-fdg21}.
The treatment variable of interest is a software patch, and the outcome of interest is the number of \emph{kills} of a team in a single match, which is a post-match statistic.
We average results over patches in the dataset.
}

\kdd{
\emph{Cannabis} is a dataset consisting of tweets about the e-cigarette Juul and cannabis-related topics~\cite{adhikari-icwsm21}.
English tweets related to the e-cigarette Juul were collected from 2016 to 2018.
For the users who mention Juul, cannabis-related tweets are collected from 2014 to 2018.
In this dataset, tweets' stances towards Juul and cannabis are crowdsourced, and a model is trained for stance detection.
In this work, we are interested in how users' stance on Juul affects their stance on cannabis.
To do this, we find users who have Juul-related tweets occurring before a cannabis-related tweet and detect their stance on Juul and cannabis using the classifier from~\cite{adhikari-icwsm21}, where features are the word embeddings for tweets using BERT~\cite{devlin-nacl19}.
The treatment variable of interest is a user's stance on Juul, in favor (1) or against (0).
The outcome variable is their stance on cannabis, in favor (1) or against (0).
Since this dataset contains only one group, we bootstrap the estimators to compute the standard deviation.
}

\begin{figure}[!t]
    \includegraphics[width=\linewidth]{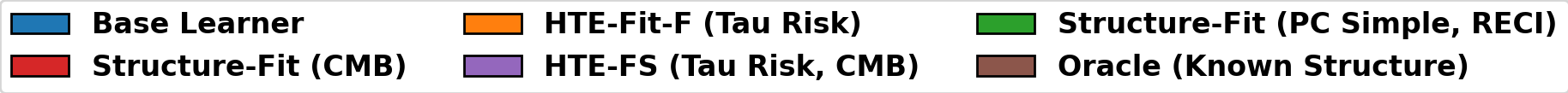}

    \includegraphics[width=\linewidth]{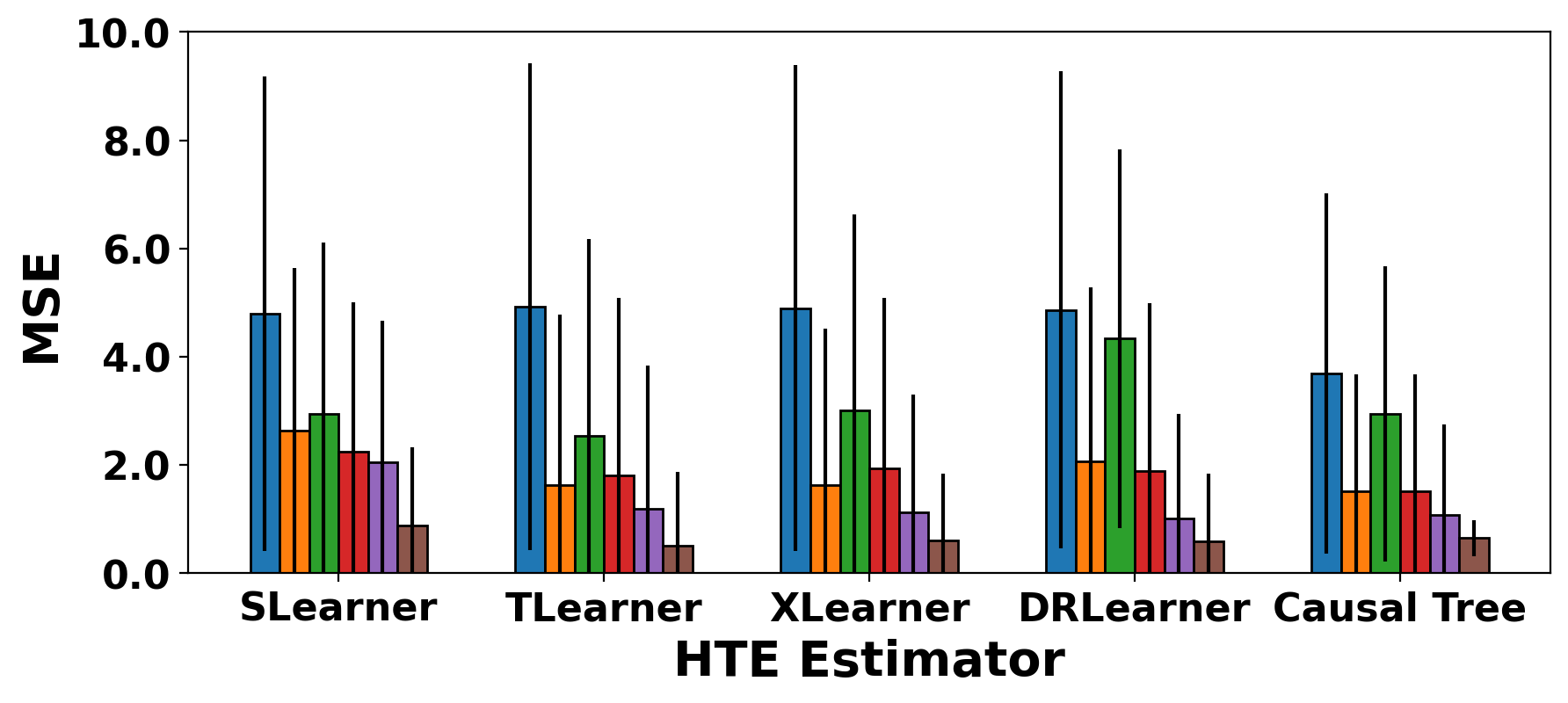}
    \caption{Average MSE for all SCMs generated using: 30 variables, confounding, mediator chain length of 2, 2 HTE inducing parents, and HTE induced by mediating parents (i.e., \( (d,\gamma,m,p_h,m_p) = (30, True, 2, 2, True) \)). We show how feature selection methods improve several HTE estimators.}\label{fig:learner_compare}
\end{figure}

\begin{table*}[!ht]
    \centering
    \resizebox*{\linewidth}{!}{
		\renewcommand{\minval}{2.0}
		\renewcommand{\maxval}{11.0}
		\begin{tabular}{ P{1.25cm} | P{3.0cm} | *{11}{M{1.75cm}} N }
			& &  Mediator Chain Length 0 &  Mediator Chain Length 1 &  Mediator Chain Length 2 &  Contains Confounding &  No Confounding &  0 HTE Parents &  1 HTE Parent &  2 HTE Parents &  10 Variables &  20 Variables &  30 Variables  \\
			\toprule
			\multirow{8}{*}{\shortstack[l]{Base\\Learners}}& SLearner & 10.58 \( \pm \) 2.25 & 10.39 \( \pm \) 2.31 & 10.38 \( \pm \) 2.19 & 10.55 \( \pm \) 2.08 & 10.36 \( \pm \) 2.38 & 10.55 \( \pm \) 2.22 & 10.08 \( \pm \) 2.38 & 10.63 \( \pm \) 2.16 & 10.56 \( \pm \) 2.41 & 10.26 \( \pm \) 2.30 & 10.55 \( \pm \) 2.09& \\[6pt] 
			& TLearner & 8.18 \( \pm \) 2.10 & 8.07 \( \pm \) 2.17 & 8.35 \( \pm \) 2.14 & 8.32 \( \pm \) 2.04 & 8.07 \( \pm \) 2.22 & 8.26 \( \pm \) 2.06 & 7.93 \( \pm \) 2.32 & 8.32 \( \pm \) 2.04 & 8.06 \( \pm \) 2.22 & 8.06 \( \pm \) 2.22 & 8.40 \( \pm \) 1.98& \\[6pt] 
			& XLearner & 8.31 \( \pm \) 2.30 & 8.37 \( \pm \) 2.28 & 8.13 \( \pm \) 2.20 & 8.28 \( \pm \) 2.14 & 8.30 \( \pm \) 2.35 & 8.24 \( \pm \) 2.24 & 8.04 \( \pm \) 2.36 & 8.51 \( \pm \) 2.18 & 8.32 \( \pm \) 2.46 & 8.15 \( \pm \) 2.29 & 8.44 \( \pm \) 2.08& \\[6pt] 
			& DRLearner & 9.64 \( \pm \) 2.53 & 9.28 \( \pm \) 2.72 & 9.73 \( \pm \) 2.73 & 9.57 \( \pm \) 2.65 & 9.42 \( \pm \) 2.71 & 9.78 \( \pm \) 2.36 & 9.14 \( \pm \) 2.90 & 9.60 \( \pm \) 2.66 & 8.94 \( \pm \) 2.66 & 9.30 \( \pm \) 2.73 & 10.11 \( \pm \) 2.53& \\[6pt] 
			& Causal Tree & 7.89 \( \pm \) 4.57 & 7.33 \( \pm \) 4.23 & 7.61 \( \pm \) 4.30 & 8.04 \( \pm \) 4.48 & 7.18 \( \pm \) 4.21 & 7.67 \( \pm \) 4.59 & 8.02 \( \pm \) 4.22 & 7.08 \( \pm \) 4.25 & 6.71 \( \pm \) 4.30 & 8.06 \( \pm \) 4.31 & 7.45 \( \pm \) 4.32& \\[6pt] 
			& Causal Forest & 6.92 \( \pm \) 3.49 & 8.14 \( \pm \) 3.63 & 6.88 \( \pm \) 3.53 & 6.90 \( \pm \) 3.51 & 7.93 \( \pm \) 3.64 & 6.67 \( \pm \) 3.39 & 7.15 \( \pm \) 3.65 & 8.26 \( \pm \) 3.59 & 8.02 \( \pm \) 3.49 & 7.07 \( \pm \) 3.65 & 7.70 \( \pm \) 3.63& \\[6pt] 
			& BNN & 7.28 \( \pm \) 3.07 & 7.65 \( \pm \) 2.98 & 7.40 \( \pm \) 3.20 & 7.49 \( \pm \) 3.18 & 7.49 \( \pm \) 2.98 & 7.72 \( \pm \) 3.10 & 7.65 \( \pm \) 3.13 & 7.25 \( \pm \) 2.98 & 7.75 \( \pm \) 2.92 & 7.61 \( \pm \) 3.17 & 7.18 \( \pm \) 3.01& \\[6pt] 
			& TARNet & 8.48 \( \pm \) 3.08 & 9.01 \( \pm \) 3.07 & 8.40 \( \pm \) 3.34 & 8.44 \( \pm \) 3.28 & 8.91 \( \pm \) 3.06 & 8.97 \( \pm \) 2.92 & 8.54 \( \pm \) 3.25 & 8.71 \( \pm \) 3.22 & 9.19 \( \pm \) 3.02 & 8.58 \( \pm \) 3.20 & 8.57 \( \pm \) 3.18& \\[6pt] 
			\midrule
			\multirow{3}{*}{\shortstack[l]{Baseline\\Feature\\Selection\\with\\TLearner}}& PC Simple (Parents) & 6.51 \( \pm \) 3.01 & 6.50 \( \pm \) 3.52 & 7.29 \( \pm \) 3.69 & 7.11 \( \pm \) 3.23 & 6.45 \( \pm \) 3.26 & 7.18 \( \pm \) 3.03 & 6.80 \( \pm \) 3.56 & 6.92 \( \pm \) 2.86 & 6.74 \( \pm \) 3.66 & 7.41 \( \pm \) 3.33 & 6.61 \( \pm \) 3.38& \\[6pt] 
			& FSL (O-set) & 4.86 \( \pm \) 2.99 & 4.37 \( \pm \) 2.85 & 4.86 \( \pm \) 2.98 & 4.73 \( \pm \) 2.88 & 4.54 \( \pm \) 2.97 & 4.52 \( \pm \) 2.93 & 4.89 \( \pm \) 3.09 & 4.45 \( \pm \) 2.78 & 4.25 \( \pm \) 3.01 & 4.81 \( \pm \) 3.02 & 4.64 \( \pm \) 2.73& \\[6pt] 
			& FSL (Valid) & 4.82 \( \pm \) 3.14 & 4.46 \( \pm \) 2.95 & 5.04 \( \pm \) 3.22 & 4.81 \( \pm \) 3.08 & 4.63 \( \pm \) 3.21 & 4.49 \( \pm \) 3.05 & 5.04 \( \pm \) 3.35 & 4.39 \( \pm \) 3.03 & 4.18 \( \pm \) 3.19 & 4.72 \( \pm \) 3.20 & 4.82 \( \pm \) 3.00& \\[6pt] 
			\midrule
			\multirow{4}{*}{\shortstack[l]{Our\\Feature\\Selection\\with\\TLearner}}& \globalmethodname{} (Tau Risk) & 4.32 \( \pm \) 1.90 & 4.33 \( \pm \) 1.76 & 4.40 \( \pm \) 2.05 & 4.30 \( \pm \) 1.98 & 4.37 \( \pm \) 1.79 & 4.38 \( \pm \) 1.92 & 4.24 \( \pm \) 1.91 & 4.43 \( \pm \) 1.83 & 4.63 \( \pm \) 1.92 & 4.22 \( \pm \) 1.87 & 4.33 \( \pm \) 1.82& \\[6pt] 
			& \localmethodname{} (PC, RECI) & 5.82 \( \pm \) 3.18 & 5.74 \( \pm \) 3.23 & 5.99 \( \pm \) 3.37 & 5.69 \( \pm \) 3.18 & 5.93 \( \pm \) 3.31 & 5.72 \( \pm \) 3.19 & 6.28 \( \pm \) 3.58 & 5.53 \( \pm \) 2.98 & 5.80 \( \pm \) 3.23 & 6.11 \( \pm \) 3.45 & 5.48 \( \pm \) 2.99& \\[6pt] 
			& \localmethodname{} (CMB) & 3.84 \( \pm \) 3.21 & 3.51 \( \pm \) 2.92 & 3.99 \( \pm \) 3.03 & 3.59 \( \pm \) 3.10 &\bftab 3.62 \( \pm \) 3.09 & 3.72 \( \pm \) 2.96 & 4.12 \( \pm \) 3.31 & 3.51 \( \pm \) 2.80 &\bftab 3.67 \( \pm \) 3.10 & 4.01 \( \pm \) 3.05 & 3.53 \( \pm \) 2.75& \\[6pt] 
			& \combined{} (Tau, CMB) &\bftab 3.67 \( \pm \) 1.96 &\bftab 3.30 \( \pm \) 2.13 &\bftab 3.69 \( \pm \) 2.32 &\bftab 3.56 \( \pm \) 2.19 & 3.69 \( \pm \) 2.53 &\bftab 3.26 \( \pm \) 2.33 &\bftab 3.94 \( \pm \) 2.73 &\bftab 3.47 \( \pm \) 2.40 & 3.73 \( \pm \) 2.59 &\bftab 3.68 \( \pm \) 2.61 &\bftab 3.30 \( \pm \) 1.89& \\[6pt] 
			\midrule
			\multirow{1}{*}{\shortstack[l]{Oracle\\with\\TLearner}}& Known Structure & 2.82 \( \pm \) 3.44 & 2.42 \( \pm \) 3.28 & 2.98 \( \pm \) 3.45 & 2.79 \( \pm \) 3.34 & 2.58 \( \pm \) 3.40 & 2.62 \( \pm \) 3.21 & 3.14 \( \pm \) 3.64 & 2.33 \( \pm \) 3.21 & 2.88 \( \pm \) 3.44 & 2.88 \( \pm \) 3.46 & 2.26 \( \pm \) 3.17 \\[6pt] 
		\end{tabular}		
		}
	\vspace{1em}
    \caption{Average rank for different HTE estimation methods under varying causal structures. 
    Each column shows a fixed parameter value, and averaged ranks over all other settings.
    The top section shows HTE estimators without feature selection.
    The second section of the table shows baseline causal feature selection methods, using TLearner as the base HTE estimator. 
	The third section shows our proposed causal feature selection methods.
    The bottom section shows the Oracle. 
    Bolded results indicate best average rank (ignoring Oracle estimation).}\label{tab:mse_rankings}
\end{table*}

\subsection{Evaluation}

We use the mean squared error (MSE) for evaluating HTE estimations since the true causal effect is known with synthetic and semi-synthetic data: $\text{MSE} = \frac{1}{N} \sum_{i=1}^N (\tau(\mathbf{X}) - \hat{\tau}(\features_i))^2$.
For each dataset, we split the data into 80/20 training and test sets.
Since different SCMs will result in different MSEs, we first rank all estimation methods based on their MSE for a particular SCM.\@
Then we average the rank of each method across SCMs and report on the average rank. 
To study their performance under specific settings, we also report on average ranks for subsets of parameter settings.
For real-world datasets, we report results using $\tau$-risk~\cite{nie-rlearner17,schuler-model18}, which we defined in eq~\eqref{eq:tau_risk}.
Similar to MSE, we report rankings based on $\tau$-risk.

\kdd{
In addition, for datasets in which we can decide which variables \emph{post-treatment} variables, we compute the percentage of included by each feature selection method, which we denote as \emph{inclusion error (IE)}.
Given the \emph{post-treatment} set, $\mathcal{F}$, and the predicted causal feature set $\mathbf{X}^{(\ell)}$, we compute the inclusion error as:
\begin{equation}
    \text{IE} = \frac{| \mathbf{X}^{(\ell)} \cap \mathcal{F} |}{|\mathcal{F}|}.
\end{equation}
Some variables occur after treatment for real-world datasets, such as other post-match statistics in LoL and tweets related to cannabis after the initial cannabis-related tweet in the Cannabis dataset.
}

\subsection{Evaluation on synthetic data}

\kdd{
We first present the results on synthetic datasets.
We first show how our feature selection method improves multiple base learners.
Then we show how feature selection improves base learners across multiple structural configurations.
}

\subsubsection{Feature selection helps all base learners}

Figure~\ref{fig:learner_compare} shows the MSE comparisons between several HTE estimators and feature selection methods on SCMs which are generated using: 30 variables, confounding, mediator chain length of 2, 2 HTE inducing parents, and HTE induced by mediating parents (i.e., \( (d,\gamma,m,p_h,m_p) = (30, True, 2, 2, True) \)).
We average MSE over all other parameters (i.e., \( p_e, \sigma, \rho \)).
\kdd{
We show the best performing \localmethodname{}, \globalforward{}, \combined{}.
This figure shows that adding feature selection improves MSE for all HTE estimators.
In addition, feature selection reduces the error bars for all base learners.
In general, \combined{} (Tau risk, CMB) performs the best, compared to using \globalforward{} or \localmethodname{} alone.
\localmethodname{} with CMB performs better than \localmethodname{} with PC Simple and RECI, although the latter still reduces error overall compared to using the base learner on all features.
}

\subsubsection{Feature selection improves base learners for different types of structures}
\kdd{
Table~\ref{tab:mse_rankings} ranks the feature selection methods using TLearner as the base learner vs.\ several other HTE estimators (without using feature selection) under a range of structural scenarios.
}
\kdd{
We choose TLearner as the base learner since it is a relatively simple model and performs consistently in all settings.
}
One experimental parameter that affects the causal structure is fixed in each column while averaging over the other parameters.
The top section shows the average rank for base learners without feature selection, the second section shows baseline feature selection methods, the third section shows our feature selection methods, and the bottom section shows the Oracle method.
Bolded results indicate the best rank in the column (ignoring the Oracle estimator).
The large variety of underlying structures explains the high variance in rank.

\kdd{
The table shows that our feature selection methods improve base learners for different structures, perform better than baseline feature selection methods, and perform better than other base learners without feature selection.
\combined{} performs the best overall.
In two settings, with no confounding and a low number of variables, \localmethodname{} with CMB performs the best.
While our methods perform better than baseline methods, baseline methods still improve base learners, compared to no feature selection.
}

\begin{figure}[!t]
    \includegraphics[width=\linewidth]{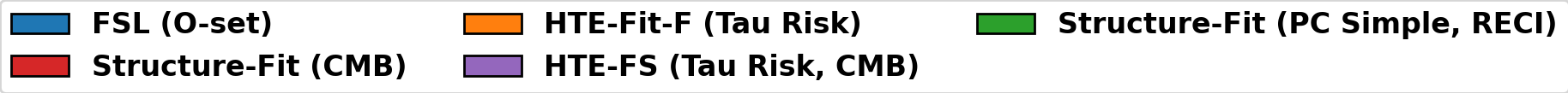}

    \includegraphics[width=0.8\linewidth]{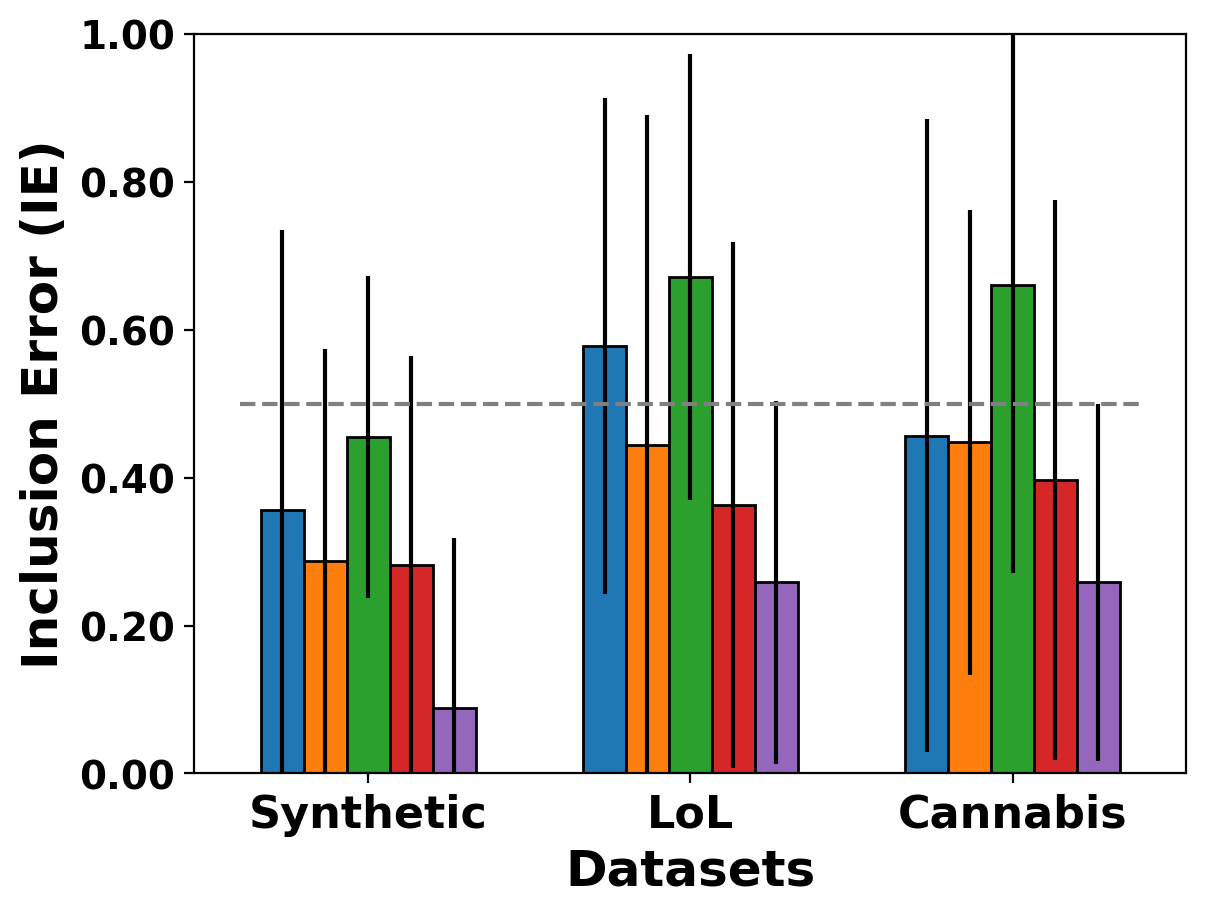}
    \caption{How often feature selection methods include descendants of treatment (lower is better). The grey line denotes 50\%, or selected at random.}\label{fig:wrong_variables}
\end{figure}

\subsubsection{Feature selection reduces inclusion of wrong variables}
\kdd{
In addition to performance in terms of MSE, we investigate the \emph{inclusion error (IE)} of feature selection methods.
Figure~\ref{fig:wrong_variables} shows the inclusion error for each feature selection method, averaged over all experiments, where lower is better.
The grey line represents the case of randomly selected features.
The first set of bars shows the average percentage of overall synthetic datasets.
All feature selection methods do better than random feature selection.
\combined{} has the lowest chance to include wrong variables, at about 8\% inclusion rate.
This result supports the results shown in Table~\ref{tab:mse_rankings}, that \combined{} has the lowest error, followed by \localmethodname{} with CMB and \globalforward{}.
}

\subsection{Evaluation on IHDP dataset}

\begin{table}[!t]
    \centering
    \resizebox*{1.0\linewidth}{!}{
		\renewcommand{\minval}{2.0}
		\renewcommand{\maxval}{11.0}
		\begin{tabular}{ P{1.1cm} | P{3.0cm} | *{4}{M{1.5cm}} N }
			& &  MSE Rank &  $\tau$-risk Rank  & LoL $\tau$-risk Rank & Cannabis $\tau$-risk Rank \\
			\toprule
			\multirow{8}{*}{\shortstack[l]{Base\\Learners}}& SLearner & 7.82 \( \pm \) 2.07 & 9.01 \( \pm \) 2.31 & 12.02 $ \pm $ 0.51 & 14.06 $ \pm $ 0.80 & \\[10pt]
			& TLearner & 4.65 \( \pm \) 1.63 & 5.16 \( \pm \) 1.79 & 4.97 $ \pm $ 0.50 & 7.86 $ \pm $ 1.03 & \\[10pt]
			& XLearner & 5.03 \( \pm \) 1.84 & 5.36 \( \pm \) 1.70 & 4.72 $ \pm $ 0.36 & 6.22 $ \pm $ 0.97 & \\[10pt]
			& DRLearner & 9.29 \( \pm \) 2.33 & 9.44 \( \pm \) 2.05 & 10.74 $ \pm $ 0.45 & 14.58 $ \pm $ 0.58 & \\[10pt]
			& Causal Tree & 8.84 \( \pm \) 2.51 & 9.14 \( \pm \) 3.02 & 8.48 $ \pm $ 0.58 & 7.01 $ \pm $ 0.92 & \\[10pt]
			& Causal Forest & 10.01 \( \pm \) 2.46 & 7.12 \( \pm \) 3.01 & 10.93 $ \pm $ 0.46 & 9.68 $ \pm $ 0.99 & \\[10pt]
			& BNN & 8.51 \( \pm \) 1.74 & 8.55 \( \pm \) 2.71 & 10.45 $ \pm $ 0.69 & 13.01 $ \pm $ 0.85 & \\[10pt]
			& TARNet & 9.24 \( \pm \) 1.61 & 8.08 \( \pm \) 2.79 & 9.09 $ \pm $ 0.54 & 12.02 $ \pm $ 0.82 & \\[10pt]
			\midrule
			\multirow{3}{*}{\shortstack[l]{Baseline\\Feature\\Selection\\with\\TLearner}}& PC Simple (Parents) & 4.61 \( \pm \) 1.67 & 4.19 \( \pm \) 1.82 & 6.51 $ \pm $ 0.64 & 11.07 $ \pm $ 0.76 & \\[10pt]
			& FSL (O-set) & 3.48 \( \pm \) 1.94 & 3.99 \( \pm \) 2.34 & 3.47 $ \pm $ 0.47 & 2.82 $ \pm $ 1.56 & \\[10pt]
			& FSL (Valid) & 4.23 \( \pm \) 1.70 & 5.17 \( \pm \) 1.32 & 4.44 $ \pm $ 0.37 & 5.02 $ \pm $ 0.61 & \\[10pt]
			\midrule
			\multirow{4}{*}{\shortstack[l]{Our\\Selection\\with\\TLearner}}& \globalforward{} (Tau Risk) & 3.30 \( \pm \) 1.89 & 3.84 \( \pm \) 2.37 & 3.97 $ \pm $ 0.44 & 2.76 $ \pm $ 0.88 & \\[10pt]
			& \localmethodname{} (PC, RECI) & 3.53 \( \pm \) 1.94 & 4.01 \( \pm \) 2.34 & 5.66 $ \pm $ 0.49 & 9.20 $ \pm $ 0.76 & \\[10pt]
			& \localmethodname{} (CMB) & 3.25 \( \pm \) 1.75 & 3.45 \( \pm \) 1.88 & 3.27 $ \pm $ 0.37 & 2.74 $ \pm $ 0.90 & \\[10pt]
			& \combined{} (Tau, CMB) &\bftab 2.77 \( \pm \) 1.70 &\bftab 3.37 \( \pm \) 1.47 & \bftab 2.48 $ \pm $ 0.44 & \bftab 1.95 $ \pm $ 1.27 & \\[10pt]
		\end{tabular}		
    }
    \caption{Average rank of estimation methods for the IHDP, LoL, and Cannabis datasets.}\label{tab:ihdp_rankings}
\end{table}

\kdd{
Table~\ref{tab:ihdp_rankings} shows the average rankings on the IHDP dataset using MSE and $\tau$-risk.
}
We used TLearner as the base learner and showed the same variations in feature selection as in Table~\ref{tab:mse_rankings}.
We see that all feature selection methods perform better on average than the base estimators, with \combined{} performing the best.
Since TLearner is a linear model close to the data generation process for this dataset, it already performs very well on it.
Since this dataset contains pre-treatment features only, we do not evaluate inclusion error.

\subsection{Evaluation on real-world datasets}

\kdd{
Finally, we show results on real-world datasets.
Since these datasets do not contain ground truth effects, we evaluate methods using $\tau$-risk rankings.
}

\subsubsection{LoL dataset}
\kdd{
The average rank for each method in terms of $\tau$-risk for the LoL dataset is shown in Table~\ref{tab:ihdp_rankings}.
Overall, \combined{} achieves the lowest $\tau$-risk rank among all estimators.
We see that TLearner performs the best among base learners and has a lower rank than some feature selection methods.
One reason could be that some methods are not suitable for high dimensional datasets, such as \localmethodname{} with PC Simple. 
However, \localmethodname{} with CMB performs much better than PC Simple, \globalforward{}, and \fullstructure{}.
}

\kdd{
In the LoL dataset, there are pre-match and post-match features. 
Since our outcome \emph{kills} is a post-match feature, other post-match features should not be included in the estimation, which is the \emph{post-treatment} set in this setting.
The second set of bars in Figure~\ref{fig:wrong_variables} show the inclusion error for the LoL datasets.
Overall, each feature selection method has a higher IE than in the synthetic dataset. 
\combined{}, \localmethodname{} with CMB, and \globalforward{} reduce the inclusion error rate on average, while \localmethodname{} with PC-Simple and \fullstructure{} do not reduce the inclusion error more than random selection.
}

\subsubsection{Cannabis}

\kdd{
The rank in terms of $\tau$-risk for the Cannabis dataset is shown in Table~\ref{tab:ihdp_rankings}.
Since this dataset contains only one set of data, we bootstrap each estimator.
In this dataset, \combined{} achieves the lowest $\tau$-risk, followed by \fullstructure{} and \globalforward{}.
Here, XLearner achieves the lowest $\tau$-risk among base learners and performs close to our proposed methods.
}

\kdd{
In the Cannabis dataset, the post-treatment features are tweets after the first cannabis-related tweet.
The features that should be included are only tweets related to the stance towards Juul and the tweet related to the first cannabis tweet.
Figure~\ref{fig:wrong_variables} shows the inclusion error for the Cannabis dataset in the rightmost set of bars.
In most cases, our feature selection methods reduce the number of wrong variables included, except \localmethodname{} with PC-Simple.
}

\section{Conclusion}
In this work, we addressed the problem of improving data-driven heterogeneous treatment effect (HTE) estimation when the underlying structural causal model (SCM) is unknown.
We developed \combined{} for causal feature selection for HTE estimation which consists of two components: \globalmethodname{} and \localmethodname{}.
We show that \combined{} reduces HTE estimation error of existing data-driven methods under a variety of causal model assumptions when compared to using their vanilla versions with all available features.
Our research suggests several interesting avenues of future research, including %
the development of causal feature selection algorithms in the presence of latent variables and the consideration of the full do-calculus, not just backdoor adjustment.

\section{Acknowledgments}

This material is based on research sponsored in part by NSF under grant No.\@ 2047899, DARPA under contract number HR001121C0168, and an Adobe Data Science Research Award. 

\bibliographystyle{ACM-Reference-Format}
\bibliography{references.bib}

\section{Appendix}

Here we provide additional information for \Combined{} and the synthetic data generation.

\subsection{\Globalmethodname{}}\label{sec:hte_fit_desc}

Here we present pseudocode for the two \Globalmethodname{} algorithms.
The \Globalforward{} algorithm is shown in Algorithm~\ref{alg:forward_selection}. Lines 1-8 first find the one feature that returns the largest score (computed by $R$) and adds it to \( \mathbf{X}^{(g)} \).
Then, for each remaining feature \( X^{(j)} \), we compute the score when combining \( X^{(j)} \) to \( \mathbf{X}^{(g)} \), and concatenate the feature that returns the best score.
This repeats until the score does not change, or there are no more features to add.

\Globalbackward{} works similarly to \globalforward{} and the pseudocode is shown in Algorithm~\ref{alg:backward_selection}.
We start by computing the score using all features in Line 1.
In Lines 9-15, we iteratively add features to \( \mathbf{K} \), which are features that when removed improve the score.
The final set returned is the whole feature set without \( \mathbf{K} \).

\begin{algorithm}[ht]
    \caption{\Globalforward}\label{alg:forward_selection}
    \begin{algorithmic}[1]
        \Require A heuristic evaluation metric $R$ and features $\features$
        \Ensure An adjustment set $\mathbf{X}^{(g)}$.

        \State bestScore = $\infty$
        \State bestFeature = None
        \For{each $X^{(j)} \in \features$}
            \State Compute tempScore $= R(X^{(j)})$
            \If{tempScore $<$ bestScore}
                \State bestFeature = $X^{(j)}$
                \State bestScore = tempScore
            \EndIf
        \EndFor
        \State $\mathbf{X}^{(g)}$ = (bestFeature)
        
        \While{bestScore changes}
            \For{each $X^{(j)} \in \features \setminus \mathbf{X}^{(g)}$}
            \State Compute tempScore = $R(\mathbf{X}^{(g)} \oplus X^{(j)})$
            \If{tempScore $<$ bestScore}
                \State bestFeature = $X^{(j)}$
                \State bestScore = tempScore
            \EndIf
            \EndFor
            \State $\mathbf{X}^{(g)} = \mathbf{X}^{(g)} \oplus$ (bestFeature)
        \EndWhile
        \State return $\mathbf{X}^{(g)}$
    \end{algorithmic}
\end{algorithm}

\begin{algorithm}[ht]
    \caption{\Globalbackward}\label{alg:backward_selection}
    \begin{algorithmic}[1]
        \Require A heuristic evaluation metric $R$ and features $\features$
        \Ensure An adjustment set $\mathbf{X}^{(g)}$.
        \State bestScore = $R(\features)$
        \State $\remove = ()$ // Features to remove
        \While{bestScore changes}
        \For{each $X^{(j)} \in \features$}
            \State Compute tempScore $= R(\features \setminus \remove \setminus (X^{(j)}))$
            \If{tempScore $<$ bestScore}
                \State $\remove = \remove \oplus (X^{(j)})$
                \State bestScore = tempScore
            \EndIf
        \EndFor
        \EndWhile
        \State $\mathbf{X}^{(g)} = \features \setminus \remove$
        \State return $\mathbf{X}^{(g)}$
    \end{algorithmic}
\end{algorithm}

\subsection{\Localmethodname{}}\label{sec:structure_fit_desc}

Algorithm~\ref{alg:lcfs} details \localmethodname{}.
In lines 1-2, we initialize two variables: $dn$, which are discovered nodes, $\mathbf{Z}$, a queue of nodes to traverse.
The discovered nodes are initialized to $ch(Y)$, so we do not traverse past $Y$.
In lines 3-5, we initialize a graph $\hat{G}$ which will be the local structure from $T$ to $Y$, and $ \mathbf{Z} $ is initialized to be the children of $ T $.
Lines 6-10 show the iterative process in which we consider the nodes in \( \mathbf{Z} \).
For every node \( Z \in \mathbf{Z} \), find \( pa(Z) \) and \( ch(Z) \).
$Z$ is added to the discovered nodes, is removed from $\mathbf{Z}$, and the children of $ Z $ are added to $ \mathbf{Z} $.
Then we connect $Z$ to its children.
In the end, a graph, $\hat{G}$ with causal paths from $T$ to $Y$ is constructed, and we can construct the valid set.

\begin{algorithm}[ht]
  \caption{\Localmethodname{}}\label{alg:lcfs}
  \begin{algorithmic}[1]
      \Require{Parent and children sets of $T$ and $Y$: $pa(T)$, $ch(T)$, \( pa(Y), ch(Y) \)}
      \State{Initialize \( dn = ch(Y) \)}
      \State{Initialize \( \mathbf{Z} = ch(T) \)}
      \State{$\mathbf{V} = \{T\} \cup ch(T)$}
      \State{$\mathbf{E} = \{(T,Z) \mid Z \in ch(T) \}$}
      \State{$\hat{G}=(\mathbf{V}, \mathbf{E}$)}
      \For{\( Z \in \mathbf{Z} \)}
        \State{Find \( pa(Z), ch(Z) \)}
        \State{\( dn = dn \cup \{ Z \} \)}
        \State{\( \mathbf{Z} = (\mathbf{Z} \cup ch(Z)) \setminus dn \)}
        \State{$\mathbf{V} = \mathbf{V} \cup \{Z\}$}
        \State{$\mathbf{E} = \mathbf{E} \cup \{(Z, Z_c \mid Z_c \in ch(Z)\}$}
      \EndFor{}
      \State{$forb(T, Y, \hat{G}) = de(T)$}
      \State{return \( \mathbf{V} \setminus forb(T, Y, \hat{G}) \)}
  \end{algorithmic}
\end{algorithm}

\subsection{Complexity of \combined{}}\label{sec:complexity}

The complexity of \combined{} depends on the complexity of training and prediction of the HTE estimator and the local structure learning algorithm.
Given an HTE estimator with complexity $ O(H) $ training and $ O(P) $ prediction, \globalmethodname{} takes $ O(d^2 \cdot (H + P) ) $.
If \globalmethodname{} outputs $ s $ variables and we use CMB for \localmethodname{}, then the worse case scenario happens when we find local structures for all $s$ variables: $ O(s \cdot 2^{|PC|}|U|^2|PC|) $~\cite{gao-neurips15}.

\subsection{Synthetic data generation}\label{sec:data_gen_desc}

Here, we describe how we generate synthetic datasets used in the experiments.
We first generate an adjacency matrix, \( \mathbf{A} \) over the \( d \) variables.
Assuming that the variables \( \{1, \dots, d \} \) are causally ordered.
For each pair of nodes \( i, j \), where \( i < j \) in the causal order, we sample from Bernoulli(\( p_e \)) to determine if there is an edge from \( i \) to \( j \).
With the adjacency matrix, we sample the corresponding coefficients, \( \mathbf{C} \), from Unif(\( -1, 1 \)).
Noise terms \( \epsilon_i \) are generated by sampling from a \( d \)-dimensional zero-mean Gaussian distribution with covariance matrix \( \Sigma \), where \( \Sigma_{i,i} = 1 \) and \( \Sigma_{i,j} \) = \( \sigma \).
We use \( \rho \) to tune the magnitude of the noise.

Next, we choose the treatment and outcome nodes based on two parameters, \( \gamma \) and \( m \). \( \gamma \) controls whether there is confounding in the generated dataset and \( m \) controls the length of the mediating chain (\( 0 \) means no mediator).
First, we select all pairs of nodes which have a \emph{directed} \( m \) hop path.
Then, if \( \gamma \) is True, we filter all pairs that do not contain a \emph{backdoor path}.
Otherwise we filter out pairs that contain a backdoor path.
Finally, we sample a pair of nodes from the filtered set to obtain the treatment and outcome nodes.
If no pairs of nodes satisfy the criteria for the given parameter setting, we resample the graph up to 100 times until a graph is found, or we discard the parameter setting.

A source node \( i \) (node with no parents) takes the value of the sampled noise terms, \( \epsilon_i \).
For a non-source node, \( i \), we consider several additive factors.
First is the base term from the coefficient matrix: \( C_{pa(i), i} X_{pa(i)} \), where \( C_{pa(i), i} \) are the coefficients from \( pa(i) \) to \( i \) and \( X_{pa(i)} \) are the values generated for \( pa(i) \).
Next, are the interaction terms, which are determined by the mediating chain length \(m\), number of HTE inducing parents \( p_h \), and whether there is HTE from parents of mediators \( m_p \).
For mediators, we have: \( \mathbbm{1}_{m}(i) \mathbbm{1}_{m_p} X_{pa_m(i)} X_{pa_n(i)} \), where \( \mathbbm{1}_{m} \) is an indicator for if \( i \) is a mediator, \( \mathbbm{1}_{m_p} \) is an indicator for if there is HTE induced from mediator parents, and \( pa_m \) and \( pa_n \) represent mediating and non-mediating parents.
For outcomes, the interaction is defined similarly: \( \mathbbm{1}_{y}(i) X_{pa_m(i)} X_{pa_n(i)} \), where \( \mathbbm{1}_{y}(i) \) indicates whether the node is the outcome node.
The number of HTE parents determines how many non-mediating parents (\( pa_n \)) the interaction occurs with.
A value of zero means there is no heterogeneity.
So the final data generating equation is:
\begin{align}
    X_i & = C_{pa(i), i} X_{pa(i)} 
    \nonumber \\
    & + \mathbbm{1}_{m}(i) \mathbbm{1}_{m_p} X_{pa_m(i)} X_{pa_n(i)} 
    \nonumber \\
    & + \mathbbm{1}_{y}(i) X_{pa_m(i)} X_{pa_n(i)} + \epsilon_i
\end{align}

\subsection{Code}\label{sec:code}

Code for \combined{} and the synthetic data generation is available at \href{https://github.com/edgeslab/causal_feature_selection}{https://github.com/edgeslab/causal\_feature\_selection}.

\end{document}